\documentclass[letterpaper, 10 pt, conference]{ieeeconf}  % Comment this line out if you need a4paper

\IEEEoverridecommandlockouts                              % This command is only needed if 
\usepackage{amsmath,amsfonts}
\usepackage{algorithmic}
\usepackage{array}
\usepackage{textcomp}
\usepackage{stfloats}
\usepackage{url}
\usepackage{verbatim}
\usepackage{graphicx}

\usepackage{multirow}
\usepackage{multicol}
\usepackage{hhline}
\usepackage{pbox}
\usepackage{graphbox}
\usepackage{booktabs}
\usepackage{cite}
\usepackage{color}
\usepackage[labelformat=simple]{subcaption}

\usepackage[flushleft]{threeparttable}

\def\BibTeX{{\rm B\kern-.05em{\sc i\kern-.025em b}\kern-.08em
    T\kern-.1667em\lower.7ex\hbox{E}\kern-.125emX}}
\usepackage{balance}

\title{\LARGE \bf
DEX-PHD: A Dexterous Portable Handheld Device\\for Multimodal Data Collection
}

\author{Jooyoung Hong, Noboru Myers, Obin Kwon, Yixian Yang, and Joohyung Kim%
\thanks{All authors are with the KIMLAB (Kinetic Intelligent Machine LAB), University of Illinois, Urbana-Champaign, IL 61801, USA 
        {\tt\small (Jooyoung Hong: jh97@illinois.edu, Joohyung Kim: joohyung@illinois.edu)}}%
}

\begin{document}

\maketitle
\thispagestyle{empty}
\pagestyle{empty}

%%%%%%%%%%%%%%%%%%%%%%%%%%%%%%%%%%%%%%%%%%%%%%%%%%%%%%%%%%%%%%%%%%%%%%%%%%%%%%%%

\begin{abstract}
Existing manipulation data collection methods typically offer either accurate robot-state acquisition through kinematic-chain-based teleoperation or convenient, portable operation through vision-based approaches. We present DEX-PHD, a portable handheld device that combines these advantages for dexterous manipulation by integrating a dexterous robot hand, continuous per-finger control, and multimodal logging. The system records proprioceptive information, fingertip contact data, and wrist-camera video, while the operator directly controls the robot hand through joint-level mapping. Its untethered design enables data collection across diverse environments. We demonstrate DEX-PHD on contact-rich tasks requiring coordinated multi-finger manipulation and deploy the collected demonstrations through open-loop replay on a physical humanoid robot. The results show that open-loop replay preserves the main temporal patterns of finger motion and task-relevant contact. Therefore, DEX-PHD provides a practical platform for portable, multimodal dexterous manipulation data collection.

% Collecting demonstrations for dexterous manipulation remains challenging because many existing pipelines are vision-only or rely on teleoperation interfaces that do not provide continuous, per-finger command generation for high-DoF robot hands. As a result, they often fail to capture synchronized hand motion, contact signals, and wrist trajectories needed for contact-rich manipulation. We present DEX-PHD, a portable handheld device that enables continuous, per-finger joint-level command generation while wirelessly recording synchronized multimodal data. By integrating a dexterous robot hand with a handheld interface, DEX-PHD records operator-generated robot-hand commands, measured hand states, fingertip force signals, and wrist-camera video directly in the robot-hand command space. The recorded demonstrations can be replayed on physical robot platforms. We evaluate DEX-PHD on dexterity-demanding tasks including peg-in-hole insertion, multi-object grasping, and drilling, and show that the collected demonstrations preserve the main temporal structure of finger motions and task-relevant contact patterns during replay. These results demonstrate the potential of DEX-PHD as a portable platform for multimodal dexterous hand data collection.
\end{abstract}

%%%%%%%%%%%%%%%%%%%%%%%%%%%%%%%%%%%%%%%%%%%%%%%%%%%%%%%%%%%%%%%%%%%%%%%%%%%%%%%%
\section{INTRODUCTION}
Robots have been increasingly deployed in industrial and household environments, where many tasks are repetitive, tedious, or physically demanding. As these applications expand, robust data collection becomes critical for advancing robotic manipulation capabilities. In particular, dexterous manipulation requires datasets that capture not only motion but also contact, force, and proprioceptive information \cite{11128816,sankar2025natural,huang2025dih}. However, collecting such data at scale remains challenging because many existing systems rely on controlled environments, specialized hardware, or complex calibration procedures.

% Such datasets are essential for training and validating control policies, benchmarking hardware, and ensuring reproducible progress across different robotic platforms. Yet, most data collection efforts have been restricted to controlled and constrained setups or specialized motion-capture environments, limiting both the scale and diversity of demonstrations. As a result, collecting rich manipulation data remains a significant bottleneck.

Existing methods for demonstration data collection present a trade-off between data quality and user convenience. Kinematic-chain-based teleoperation systems, including leader-follower devices and exoskeletons, can provide accurate robot-action trajectories and high-quality joint-level data. However, these systems often require expensive hardware and bulky or fixed setups, limiting portability and scalable data collection. In contrast, vision-based methods, including approaches based on egocentric video, provide convenient and accessible data collection. Nevertheless, they depend on visual pose estimation and retargeting, may require substantial computation, and make it difficult to incorporate contact and proprioceptive information.

% Existing approaches often rely on video recordings, VR/AR interfaces, or teleoperation devices. While these methods provide valuable data, they are typically constrained by vision-only estimation, tethered operation, or a lack of multimodal signals. In particular, precise hand–object interactions, tactile responses, and proprioceptive signals are difficult to capture with current systems, which are critical for dexterous manipulation. Moreover, many teleoperation devices are designed for real-time control rather than systematic dataset generation, and often suffer from bulkiness, limited portability, or kinematic mismatches.

Handheld devices offer a promising intermediate approach by combining simple setup with portable, in-the-wild data collection~\cite{sanches2023scalable, chi2024universal}. However, most existing handheld systems employ parallel grippers or tool-based interfaces, limiting their ability to capture continuous, coordinated multi-finger motion. They also provide limited access to synchronized hand-state and fingertip-contact information. For dexterous manipulation, there is therefore a need for a portable collection interface that retains the convenience of vision-based systems while providing direct, high-quality robot-hand data.

% Recently, handheld devices have emerged as a lightweight alternative for collecting human-guided demonstrations \cite{sanches2023scalable, chi2024universal}. Their simple setup and mobility enable users to collect data without relying on complex or bulky systems. In addition, their portability makes them well-suited for scalable, in-the-wild data collection across diverse environments. Yet, most prior designs have been limited to parallel grippers, restricting their ability to capture the full richness of dexterous multi-finger behaviors, and they often lack integrated sensing.

\begin{figure}[t]
\vspace*{5pt}
\centering
    \includegraphics[width=0.96\columnwidth]{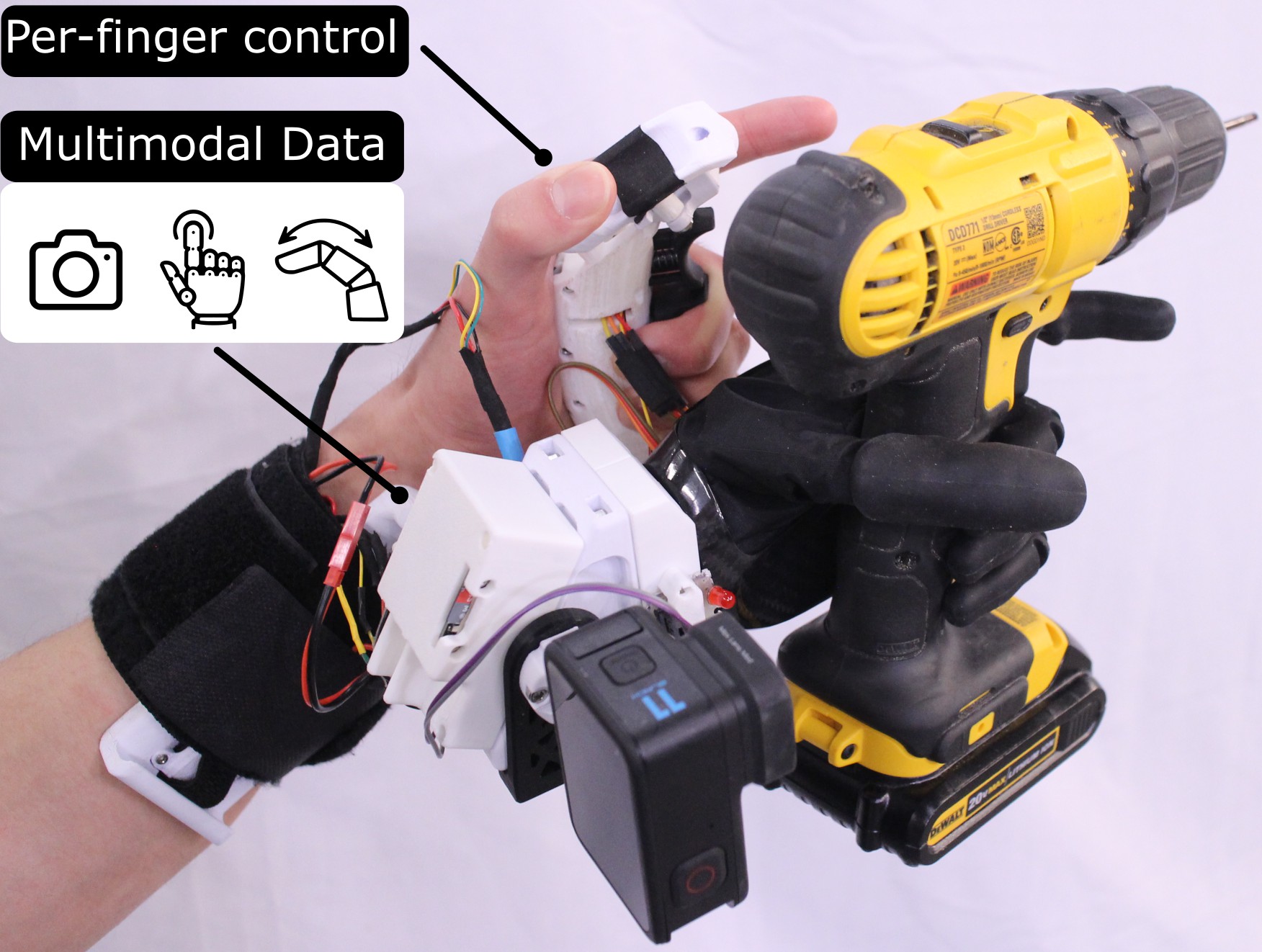}
    \caption{DEX-PHD: A dexterous portable handheld device for collecting multimodal demonstrations with a dexterous robot hand. By placing the handle alongside the robot hand, DEX-PHD enables intuitive, per-finger control and collects time-synchronized multimodal data from the dexterous robot hand and the wrist camera.}

    \label{fig:introduction}
\end{figure}

To bridge the gap between direct robot-hand data collection and portable operation, we present DEX-PHD, a portable handheld device that physically integrates a dexterous robot hand with continuous per-finger control and synchronized multimodal logging, as shown in Fig.~\ref{fig:introduction}. Continuous trigger and knob inputs are mapped directly to the corresponding robot-hand joints, enabling the operator to control the dexterous hand through a joint-level mapping. The device records hand joint states, fingertip force signals, and wrist-camera video using onboard storage. Its battery-powered, untethered configuration enables data collection across diverse environments without an external computer or fixed robot platform.

The contributions of this paper are summarized as follows:
\begin{itemize}
    \item A portable handheld device that physically integrates a dexterous robot hand and enables continuous, per-finger joint-level command generation directly in the robot-hand command space.
    \item A synchronized multimodal logging framework that records wrist-view video, robot-hand commands, measured finger states, and fingertip force signals during contact-rich manipulation.
    \item A replayability evaluation showing that demonstrations collected with DEX-PHD can be reproduced on a physical humanoid robot, with quantitative analysis of joint trajectory tracking and task-relevant contact-pattern similarity.
\end{itemize}

\section{Related Work}

% \subsection{Teleoperation Systems}
% Teleoperation systems provide a direct way to collect robot action trajectories for imitation learning \cite{qin2023anyteleop,kareer2025egomimic}. Direct joint-mapping and AR/VR-based systems further support precise bimanual teleoperation, sizable robot datasets, and cross-platform trajectory collection \cite{zhao2023learning,Wu2024GELLO,yang2024ace,iyer2024open,cheng2024opentelevision}. DEX-PHD follows the goal of collecting replayable robot trajectories, but uses a compact handheld device to record wrist trajectories and robot-hand data without requiring the full robot platform during demonstration collection.

% \subsection{Wearable and Handheld Devices}
% Wearable gloves, motion-capture systems, and hand exoskeletons capture dexterous human hand motion or provide haptic feedback outside conventional robot teleoperation setups \cite{Zhang2025DOGlove,tao2025dexwild,wang2024dexcap,fang2025dexop,Tang2026MFE,xu2025dexumi}. Handheld devices provide a portable alternative for collecting demonstrations without the full robot, using wrist-view, gripper-centric, tactile, force, or tool-based interfaces for in-the-wild and contact-rich demonstration collection \cite{zhu2025touch,chi2024universal,11128180,seo2025legato,liu2025forcemimic}. In contrast, DEX-PHD physically integrates a dexterous robot hand into the handheld device and records demonstrations directly in the robot-hand command space, enabling continuous per-finger command generation with synchronized hand states, fingertip force signals, and wrist-camera video.
\subsection{Dexterous Data Collection Interfaces}

Dexterous manipulation demonstrations can be collected through teleoperation systems, wearable interfaces, vision-based methods, and handheld devices. Teleoperation systems provide a direct way to collect robot action trajectories for imitation learning \cite{qin2023anyteleop,kareer2025egomimic}. Direct joint-mapping and AR/VR-based systems further support precise bimanual teleoperation, large-scale robot data collection, and cross-platform trajectory transfer \cite{zhao2023learning,Wu2024GELLO,yang2024ace,iyer2024open,cheng2024opentelevision}. However, these systems often require specialized hardware, calibration procedures, or a fixed robot platform during demonstration collection.

Wearable gloves, motion-capture systems, and hand exoskeletons capture dexterous human hand motion or provide haptic feedback outside conventional robot teleoperation setups \cite{Zhang2025DOGlove,tao2025dexwild,wang2024dexcap,fang2025dexop,Tang2026MFE,xu2025dexumi}. Since these systems measure human hand motion, the recorded motion generally requires retargeting before it can be applied to a robotic hand. Handheld devices provide a more portable alternative for collecting demonstrations without the full robot, using wrist-view, gripper-centric, tactile, force, or tool-based interfaces for in-the-wild and contact-rich manipulation \cite{zhu2025touch,chi2024universal,11128180,seo2025legato,liu2025forcemimic}. Nevertheless, many existing handheld systems focus on parallel grippers or tool-level interfaces and provide limited continuous control of multiple robotic fingers.

DEX-PHD integrates a dexterous robot hand into a portable handheld device, enabling direct robot-hand command generation without human-to-robot finger retargeting. It also synchronously records wrist-camera video, hand states, and fingertip force signals without requiring a fixed robot platform or external computer.

\subsection{Multimodal Data Collection}
For contact-rich manipulation, visual trajectories alone are often insufficient because task success depends on contact timing, compliance, and interaction forces. Recent studies show that tactile, force, proprioceptive, and visuo-tactile feedback can improve policy learning for insertion, assembly, and compliant manipulation \cite{xue2025reactive,huang2025vtrefine,chen2026multimodal}. Demonstration-collection systems have also begun incorporating force or tactile sensing to capture interaction dynamics that are not directly observable from vision alone \cite{zhu2025touch,liu2025forcemimic}. DEX-PHD follows this direction by synchronously recording wrist-camera video, robot-hand commands, measured hand states, motor currents, and fingertip force signals directly from the dexterous hand during demonstration collection.

\section{Method}\label{sec:method}
DEX-PHD is a portable data collection device for capturing contact-rich dexterous manipulation demonstrations. The system provides the following key capabilities:
\begin{itemize}
    \item \textbf{Dexterous control}: DEX-PHD is designed to support an intuitive human-robot mapping for continuous, per-finger control of a high-DoF hand, allowing demonstrations with fine-grained articulation beyond binary grasp commands. This supports demonstrations for multi-finger manipulation tasks that require coordinated finger motions.
    
    \item \textbf{Multimodal data}: DEX-PHD enables on-device collection of synchronized multimodal streams. The high-DoF hand provides tactile and proprioceptive data, while the wrist-mounted camera captures the scene with time-aligned IMU measurements embedded in each frame. Wrist trajectories can be extracted from the recorded videos.

    \item \textbf{Portability:} DEX-PHD is a handheld, wireless system that supports data collection in diverse environments beyond the lab. Its untethered design enables rapid setup and demonstration collection without requiring an external PC or a fixed robot platform.
\end{itemize}

\subsection{Hardware System}

\begin{figure}[t]
\vspace*{5pt}
\centering
\includegraphics[width=0.96\columnwidth]{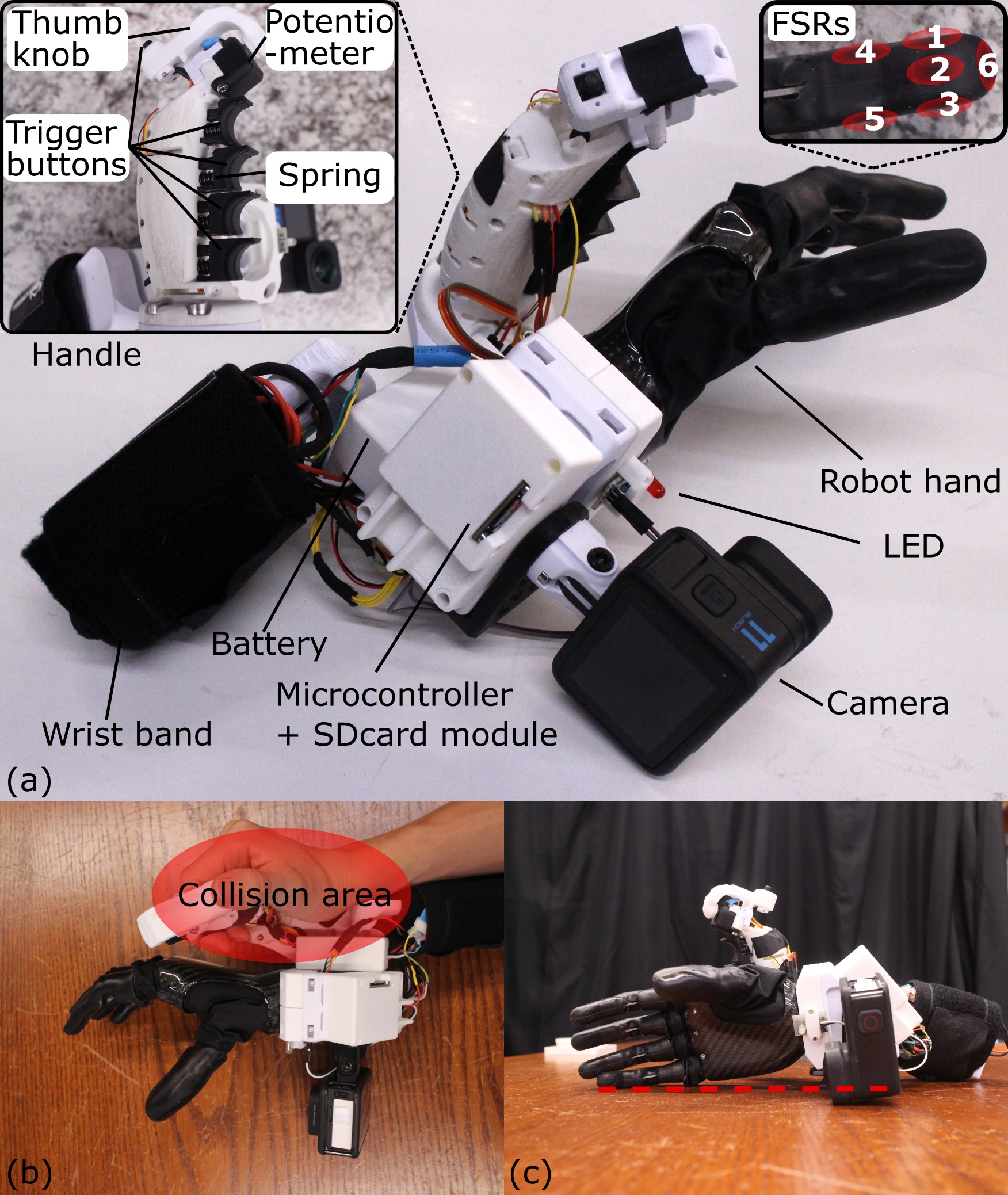}
% \caption{DEX-PHD hardware overview. (a) The robot hand can be controlled with a continuous joint mapping from analog values of the trigger buttons and a thumb knob. Tactile data can be measured from 6 FSRs on each finger. All data is logged into SD cards. (b) The robot hand could be next to a human hand by adopting a handle instead of linkage mechanisms. (c) The bottom plane of the device aligns with the one of the hand, enabling table-level reach during planar manipulation.}
\caption{DEX-PHD hardware overview. (a) Continuous joint mapping via trigger buttons and a thumb knob, with tactile sensing from six FSRs per finger and SD card logging. (b) Handle design enabling side-by-side operation with the human hand. (c) Aligned device-hand bottom planes for table-level reach during planar manipulation.}
\label{fig:hardware_system}
\end{figure}

The DEX-PHD system, shown in Fig.~\ref{fig:hardware_system}, is a portable handheld data-collection device built around a 6-DOF robotic hand (Ability Hand, Psyonic~\cite{Psyonic}). The system is designed to allow a human operator to physically carry and manipulate the robotic hand while recording multimodal demonstrations from a hand-centered perspective. A custom handle is mounted alongside the dorsal side of the robotic hand, forming a compact handheld unit that can be operated without external teleoperation hardware or a fixed robot platform.

The handle integrates independent flexion triggers for the index, middle, ring, and pinkie fingers, a thumb flexion trigger, and a dedicated thumb rotation knob. 
A potentiometer measures the thumb rotation knob, while each flexion trigger is instrumented with a Hall sensor-magnet pair. Low-stiffness return springs restore the triggers to their neutral positions after release without discretizing the continuous Hall-sensor input. 
% A potentiometer measures the thumb rotation knob, while each flexion trigger is instrumented with a Hall sensor--magnet pair. Internal springs are embedded in the trigger mechanisms so that the triggers passively return to their neutral positions after release, improving repeatability across demonstrations. 
These components provide the physical input channels used for the joint-level command mapping described in Section~\ref{sec:joint_level_mapping}.

Beyond the input mechanisms, the handle geometry is intentionally refined to improve usability in contact-rich, real-world manipulation. The handle is mounted close to the dorsal side of the robotic hand to reduce the collision footprint of the device, thereby lowering the likelihood of unintended impacts with the environment during operation. In addition, the bottom surface of the handle is aligned with the robot palm. This co-planar alignment is important for planar, table-top manipulation because it preserves clearance comparable to a human hand and allows the robotic hand to approach and grasp table-level objects without the handle contacting the surface first. Together, these structural choices make the device more natural to use during demonstration collection.

The onboard electronics consist of a microcontroller, a microSD logging module, a battery, and an LED. During data collection, the system records hand proprioceptive states, user command inputs, and tactile signals from 30 FSRs, with six FSRs per finger, to a microSD card at 100 Hz. A wrist-mounted camera captures the operational scene for SLAM-based trajectory estimation at 60 fps and is synchronized with the hand data using the onboard LED. The total system weight is 1.1 kg, with a practical runtime of 1.5 hours to maintain battery health. Detailed specifications are summarized in Table~\ref{table:specification}.

\begin{table}
\vspace*{5pt}
\centering
\caption{Specification of DEX-PHD}
\label{table:specification}
\begin{tabular}{|c|c|}
\hline
%  & \shortstack{Passive\\jamming} & \shortstack{Active\\jamming} & \shortstack{Non-\\jamming}\\
% \hline

Microcontroller & Teensy 4.0 \\
\hline
Robot hand & Ability Hand (Psyonic) \\
\hline
Control fingers & Index, middle, ring, pinkie, thumb, thumb rotation \\
\hline
Finger range  & $0 \sim 110^{\circ}$ \\
\hline
Wrist camera & Hero 11 Black (GoPro) \\
\hline

Data format & CSV (Hand) \& MP4 (Camera) \\
\hline

Weight & 1.11 kg (0.66 kg without the hand)\\
\hline
Logging rate & 100 Hz (Hand) \& 60 fps (Camera) \\
\hline
Runtime & 1.5 hours \\

\hline
\end{tabular}
\end{table}

\subsection{Joint-Level Mapping}\label{sec:joint_level_mapping}
% % DEX-PHD converts the operator's continuous trigger and knob inputs into joint-level commands for the Ability Hand. 
% Using the physical input channels described above, DEX-PHD converts the operator's trigger and knob inputs into joint-level commands for the Ability Hand.
% Each flexion trigger is measured by a Hall sensor--magnet pair, and the thumb rotation knob is measured by a potentiometer. Before each data-collection session, the minimum and maximum values of each input channel are calibrated from the neutral and fully actuated configurations. The measured input is then normalized between these calibrated bounds and linearly mapped to the corresponding motion range of the robotic hand.
Using the physical input channels described above, DEX-PHD converts the operator's continuous trigger and knob inputs into joint-level commands for the robot hand. Before each data-collection session, the minimum and maximum values of each input channel are calibrated from the neutral and fully actuated configurations. Each measured input is then normalized between these calibrated bounds and linearly mapped to the corresponding motion range of the robotic hand.

For the four fingers, the index, middle, ring, and pinkie triggers independently command the flexion of the corresponding digits. The thumb flexion trigger commands thumb closing, while the thumb knob commands thumb rotation. This one-to-one mapping provides an intuitive command space that allows the operator to independently coordinate the fingers for multi-finger reconfiguration, partial grasping, tool use, and sequential object release.

During data collection, the generated joint-level commands are sent to the robotic hand, and the corresponding command inputs, motor positions, currents, and fingertip force values are logged. This logging procedure records both the commanded hand motion and the measured hand response, enabling the recorded robot-hand command trajectories to be used for later replay and analysis.

We also characterized the responsiveness of the joint-level command pathway. The hand control and logging loop runs at 100 Hz, corresponding to a 10 ms update period. The end-to-end hand response delay was estimated by comparing the onset of the Hall-sensor trigger input with the onset of the corresponding measured joint-position response. 
Within the 10~ms temporal resolution of the onboard logger, typical finger responses were observed in the same control cycle as the corresponding input events, indicating that the input-to-response delay is below or close to the measurement resolution of the 100~Hz logging system.
% indicating a response delay within one 10 ms update period.

\subsection{Trajectory Generation}
This paper adopted the visual SLAM pipeline similar to the UMI gripper to obtain portable, egocentric trajectories\cite{chi2024universal}. The wrist camera records RGB and IMU data, from which we estimate the device pose over time. Before each task, the user briefly scans a base fiducial and the scene to establish a local base frame $\{B\}$ and keypoints for localization. 

% Many prior high-DOF teleoperation systems rely on worn exoskeletons, which can introduce kinematic mismatch and workspace constraints. In contrast, our handle is colocated with the robot hand, enabling direct and intuitive mapping without donning an exoskeleton, while avoiding embodiment gaps during data collection. 

$\mathbf{T}^{i}_{j}(t)\in SE(3)$, where $i$ and $j$ are frames, denotes the homogeneous transformation matrix at time step $t$. The end effector trajectory from the world frame, $\mathbf{T}^{E}_{W}(t)$, can be derived from the generated trajectory, $\mathbf{T}^{E}_{B}(t)$, as below:
\[
\mathbf{T}^{E}_{W}(t)\ = \mathbf{T}^{E}_{B}(t) \mathbf{T}^{B}_{W}(t),
\]
where the $\{B\}$, $\{W\}$, and $\{E\}$ stand for fiducial marker frame, robot's world frame, and end effector frame. 
When scanning the environment with the base marker, the location and orientation might be different each time, so the pose of the marker from the world frame is uncertain. Also, if the trajectory is far from the world frame, the robots cannot follow the trajectories. 
Therefore, the marker to robot transformation may be replaced with an offset transform so that $\mathbf{T}^{B}_{W}(t) = \mathbf{T}_{\text{offset}}$.
% Therefore, the offset of the coordinate is needed before replay. So the offset can be used instead of $\mathbf{T}^{B}_{W}(t)$, such as $\mathbf{T}^{B}_{W}(t) = \mathbf{T}_{\text{offset}}$.

% An example generated trajectory is shown in Fig.~\ref{fig:trajectory_generation}.

% About the bimanual trajectories, the same background videos are used for both hands, so each hand shares the same base frame. Therefore, each trajectory can be mapped from the world frame, respectively.

% \subsection{Data Synchronization}
% As image data and robot hand data are saved in each local system, it is important to align the durations of both data for training. To match the timestamps, users trigger the LEDs on the wrists by clicking the thumb flexion trigger button twice within 500 ms at the starting and ending points of each episode. Then the LEDs will blink for 1 second. The synchronizing node during post-processing to extract the wrist trajectory will clip the videos before and after the LED blink.
% % as the wrist-mounted camera captures the blinking lights. 

% For bimanual tasks, precise synchronization between the two devices is crucial as they
% produce asynchronous data streams with different durations. Following the methodology
% in \cite{chi2024universal}, we utilized a rolling QR code, which is a sequence of time-encoded QR codes displayed on a screen and captured by both cameras, to map each frame’s timestamp to a unified system time. This allows for accurate offline alignment of multimodal data, ensuring temporal consistency across both hands.

\subsection{Data Synchronization}
As image data and robot hand data are saved in each local system, it is important to align the durations of both data for analysis and replay. To match the timestamps, users trigger the LEDs on the wrists by clicking the thumb flexion trigger button twice within 500 ms at the starting and ending points of each episode. Then the LEDs blink for 1 second. The synchronizing node clips the videos before and after the LEDs blink during post-processing to extract the wrist trajectory.
% as the wrist-mounted camera captures the blinking lights. 

During post-processing, the LED blink is detected at the video-frame level, and the hand-data sequence is aligned to the camera timeline. The hand data is logged at 100~Hz, corresponding to a 10~ms sampling interval, while the wrist-camera records at approximately 60~fps, corresponding to a 16.7~ms frame interval. After temporal alignment, each image frame is paired with the hand-data sample having the nearest timestamp. Therefore, the nominal hand--video synchronization resolution is on the order of tens of milliseconds, determined by the 10~ms hand-data sampling interval and the 16.7~ms camera frame interval.

For bimanual tasks, precise synchronization between the two devices is crucial as they
produce asynchronous data streams with different durations. Following the methodology
in \cite{chi2024universal}, we utilized a rolling QR code, which is a sequence of time-encoded QR codes displayed on a screen and captured by both cameras, to map each frame’s timestamp to a unified system time. This allows for accurate offline alignment of multimodal data, ensuring temporal consistency across both hands.

\section{Demonstrations}
In this section, we evaluate three aspects of DEX-PHD. First, can DEX-PHD support untethered multimodal demonstration collection across diverse physical environments? Second, does the continuous per-finger command interface enable demonstrations that require coordinated multi-finger manipulation? Finally, can the collected robot-hand commands and wrist trajectories be deployed on physical robot platforms while preserving task-relevant motion and contact patterns?
% In this section, we evaluate three aspects of DEX-PHD.
% First, can DEX-PHD capture synchronized multimodal data from a dexterous hand and extract replayable wrist trajectories from the recorded videos?
% Second, does DEX-PHD support continuous joint-space control for real-world demonstrations that require coordinated multi-finger manipulation?
% Finally, can the collected demonstrations be deployed across different robot configurations and environments?

The collected demonstrations were deployed using open-loop replay on two different robot embodiments in simulation: a humanoid robot with 7 DoFs in each arm (Unitree G1) \cite{unitree_g1} and a 7-DoF dual-arm mounted on a fixed table. In addition, the collected demonstrations were deployed on a physical humanoid robot. The same robotic hands were mounted at the wrist frame of each robot.

\subsection{Data Collection}
In each task, two data formats were saved: videos and hand data. The video contains images and IMU data, while the hand data includes positions, currents, fingertip force values, and trigger positions.
% Unlike other works, users directly teach the robot hands how to perform tasks, avoiding kinematic mismatches.
In our setup, the operator generates demonstrations directly in the robot-hand command space, reducing the need for human-to-robot finger retargeting.

\begin{figure}[t]
\vspace*{5pt}
\centering
    
    \begin{subfigure}{0.9\columnwidth}
    \centering
    \includegraphics[width=\columnwidth]{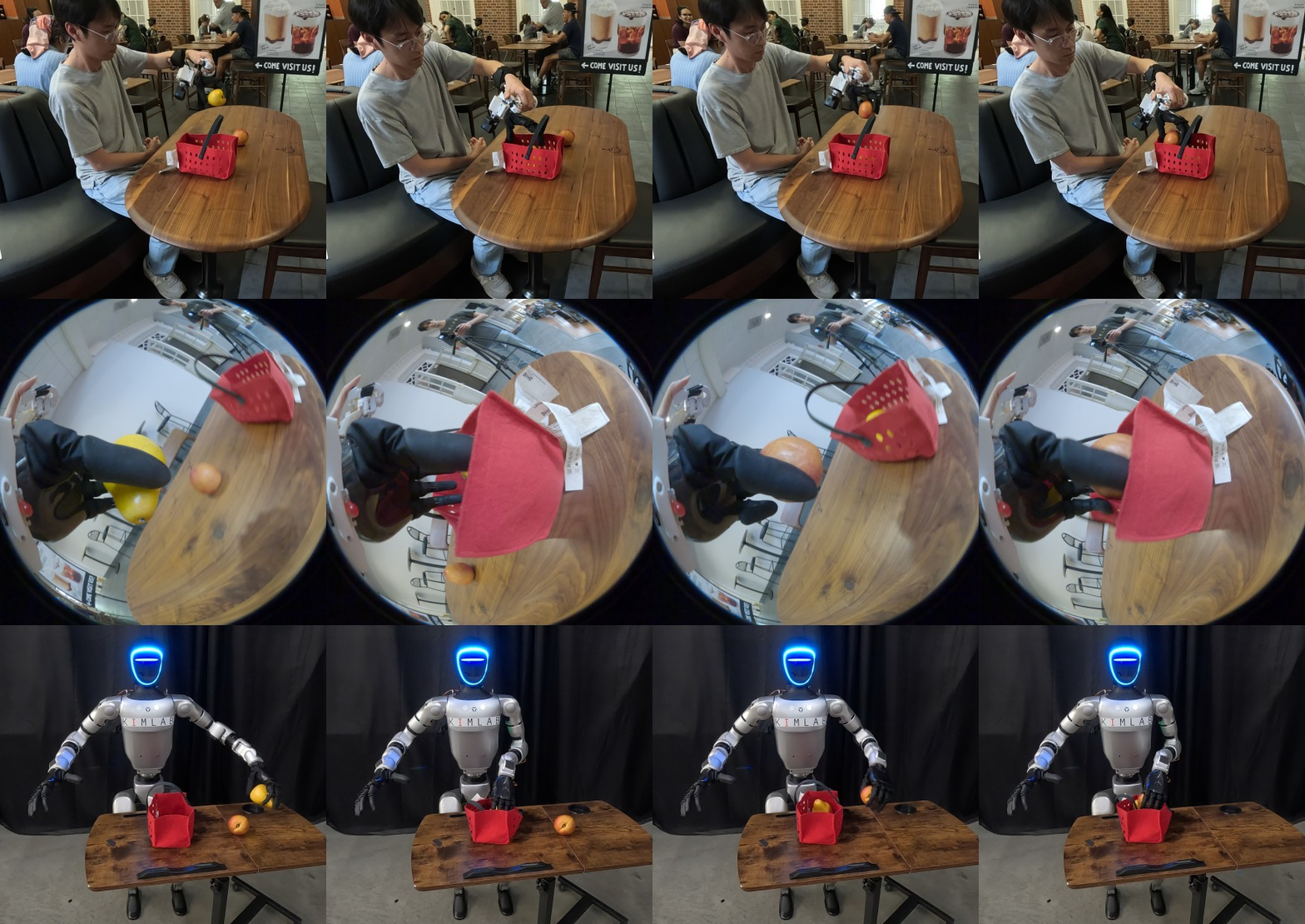}
    \caption{Pick-and-place in a cafe}
    \label{fig:union_fruit}
    \end{subfigure} 
    
    \begin{subfigure}{0.9\columnwidth}
    \centering
    \includegraphics[width=\columnwidth]{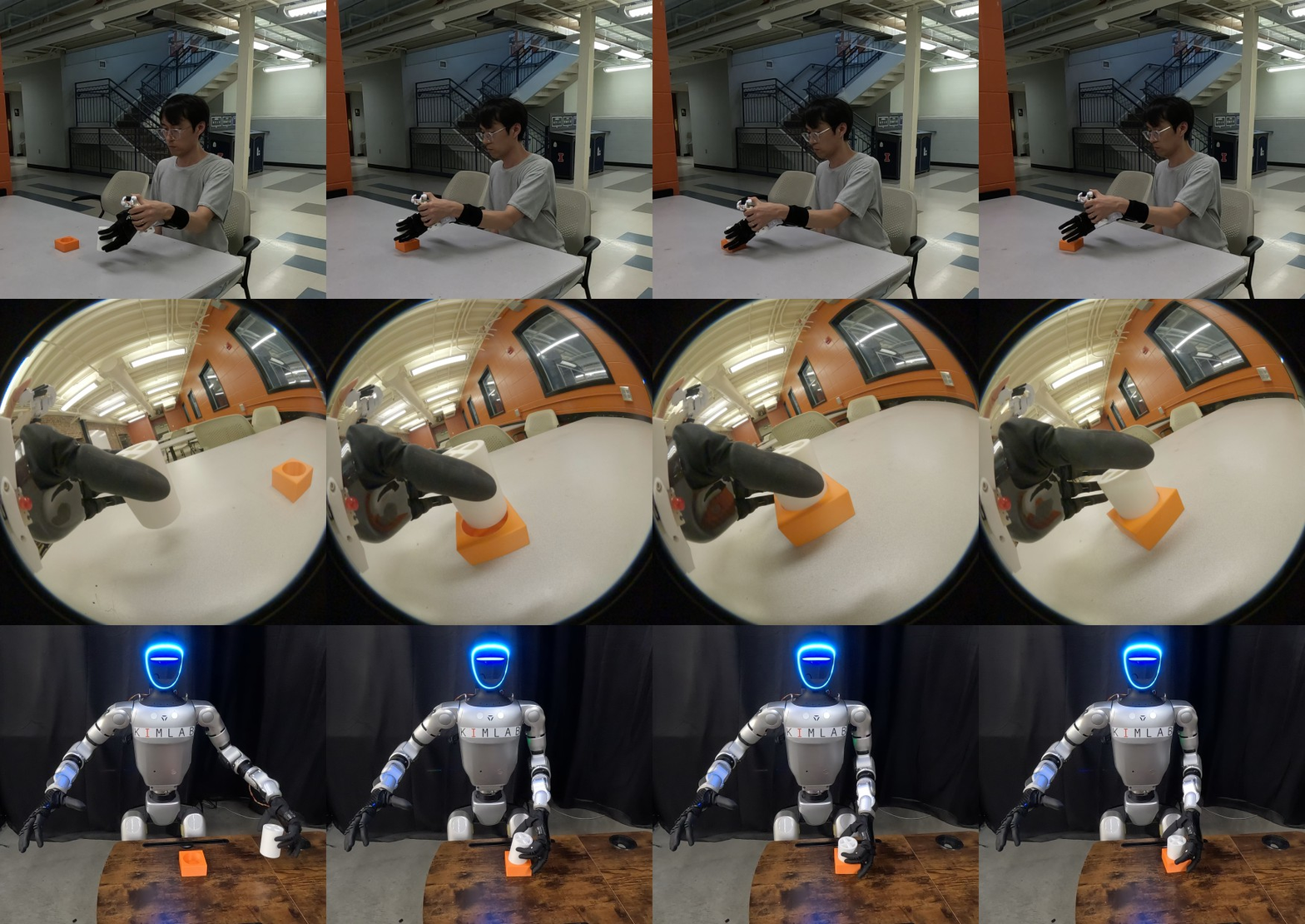}
    \caption{Peg-in-hole insertion in a building}
    \label{fig:mel_peg}
    \end{subfigure}       

    \begin{subfigure}{0.9\columnwidth}
    \centering
    \includegraphics[width=\columnwidth]{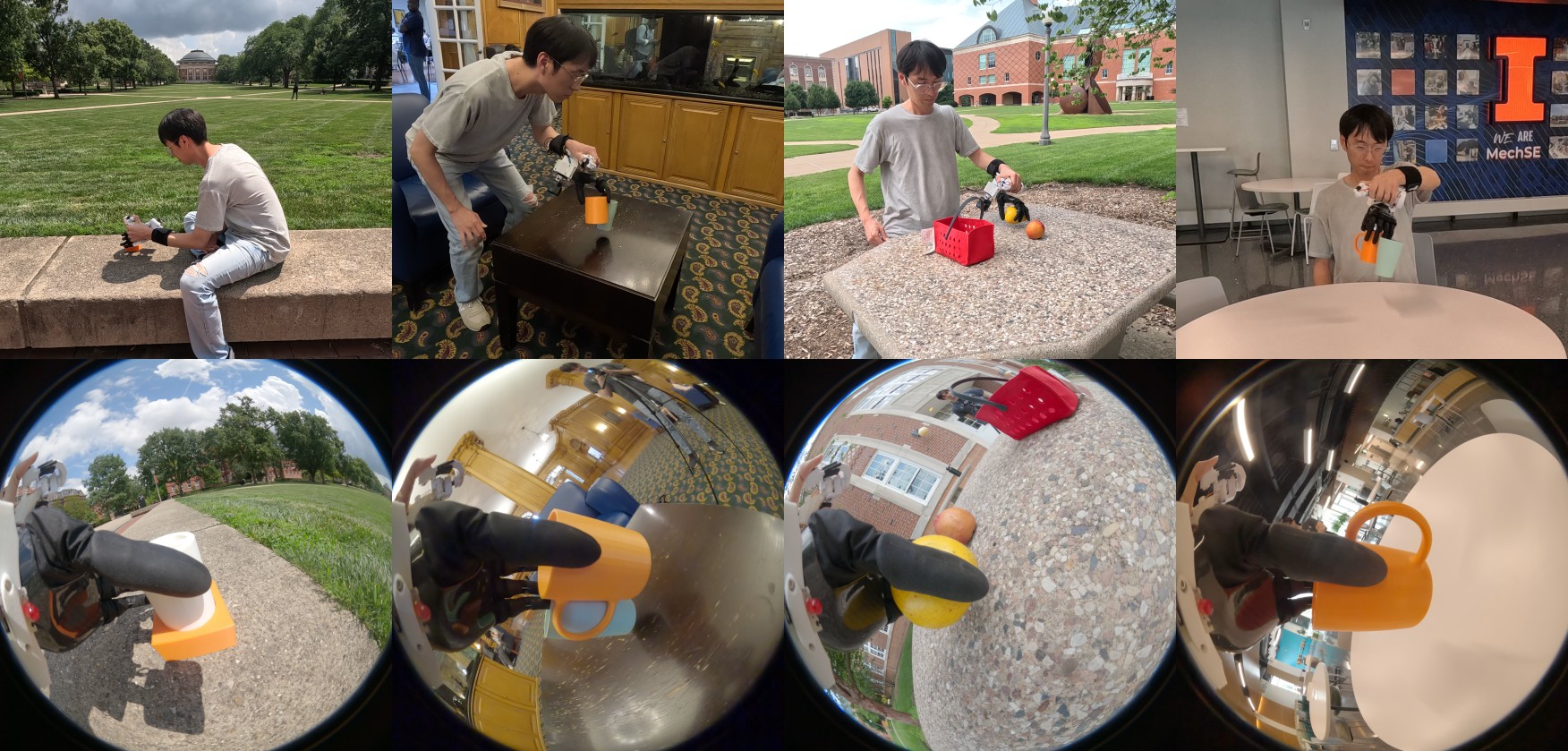}
    \caption{Data collection across diverse environments}
    \label{fig:outdoor_data}
    \end{subfigure} 
\caption{Data collected with DEX-PHD across diverse indoor and outdoor environments and subsequently replayed on a physical humanoid robot.} 
\label{fig:portability}
\end{figure}

The portability of DEX-PHD was evaluated through deployment across multiple collection environments. Its battery-powered, onboard logging architecture enabled untethered operation without an external PC or wired connection, while the device could be set up in less than one minute and calibrated in less than one minute before data collection. As shown in Fig.~\ref{fig:portability}, these characteristics allowed demonstrations to be collected in diverse indoor and outdoor environments, including a cafe, campus buildings, picnic tables, and courtyards. Demonstrations collected at these locations were subsequently deployed on a physical humanoid robot, demonstrating a portable workflow from data collection in varied environments to physical robot replay.

The following tasks were selected to evaluate whether DEX-PHD can capture multimodal data requiring coordinated finger motion and task-relevant contact interactions:

\begin{enumerate}
\item \textbf{Peg-in-hole insertion}: Peg-in-hole alignment generates frequent contact and interaction forces, making fingertip force data crucial for capturing tactile feedback during insertion. 
With a clearance of less than 0.5 mm, this task is sensitive to trajectory accuracy and contact alignment.
% With a clearance of less than 0.5 mm, this task requires high trajectory precision and robust tracking under contact.

\item \textbf{Multi-object grasping}: Sequentially grasping and placing multiple cups requires repeated reconfiguration of finger poses and stable grasp execution. This task highlights continuous per-finger control by demanding coordinated motions for reliable object acquisition and release.

\item \textbf{Drilling}: Operating a drill requires maintaining a stable power grasp while coordinating thumb rotation with trigger actuation. This task evaluates per-finger joint-space control under realistic tool-use constraints.
\end{enumerate}

\begin{figure}[t]
\vspace*{5pt}
\centering
    \begin{subfigure}{0.96\columnwidth}
    \centering
    \includegraphics[width=\columnwidth]{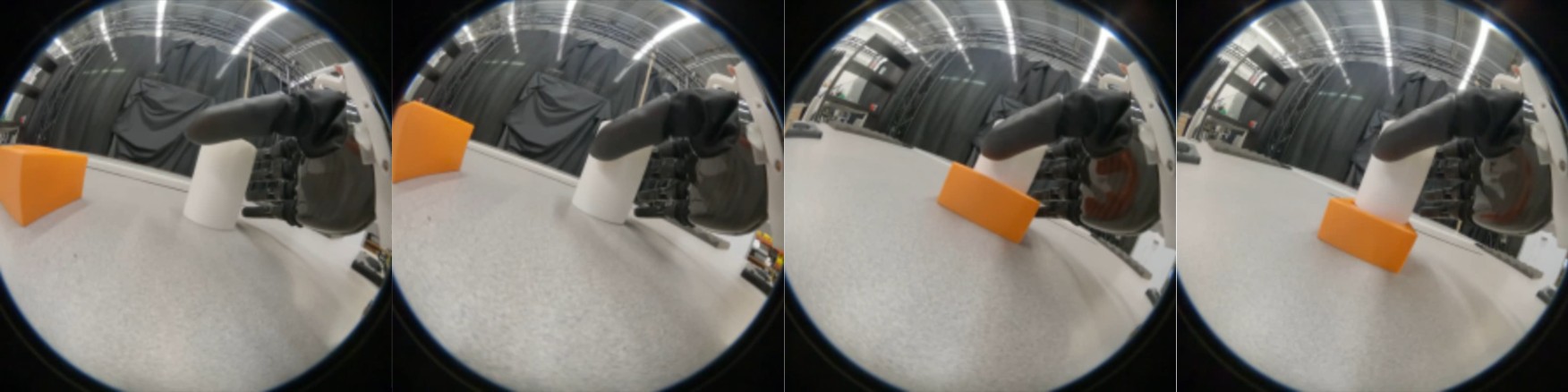}
    \caption{DEX-PHD camera view}
    \label{fig:cam_view_peg}
    \end{subfigure} 

    \begin{subfigure}{0.96\columnwidth}
    \centering
    \includegraphics[width=\columnwidth]{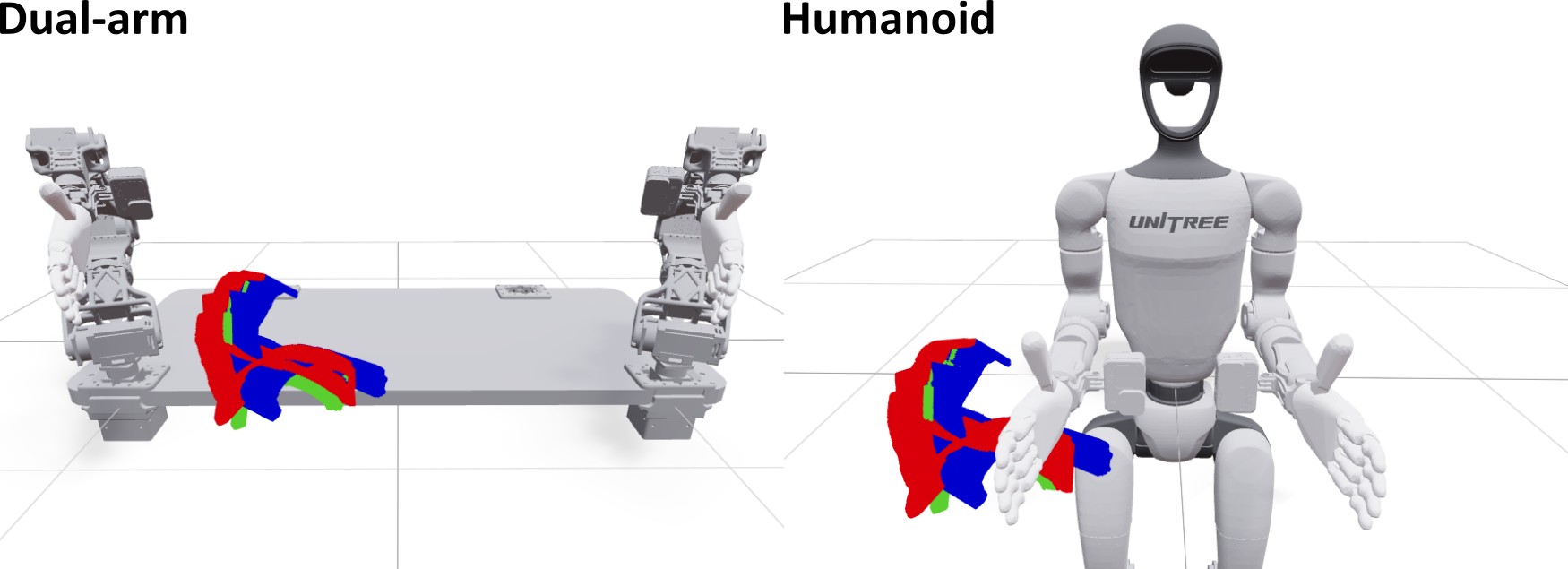}
    \caption{Trajectories generated using SLAM pipeline}
    \label{fig:trajectory_generation}
    \end{subfigure} 

    \begin{subfigure}{0.96\columnwidth}
    \centering
    \includegraphics[width=\columnwidth]{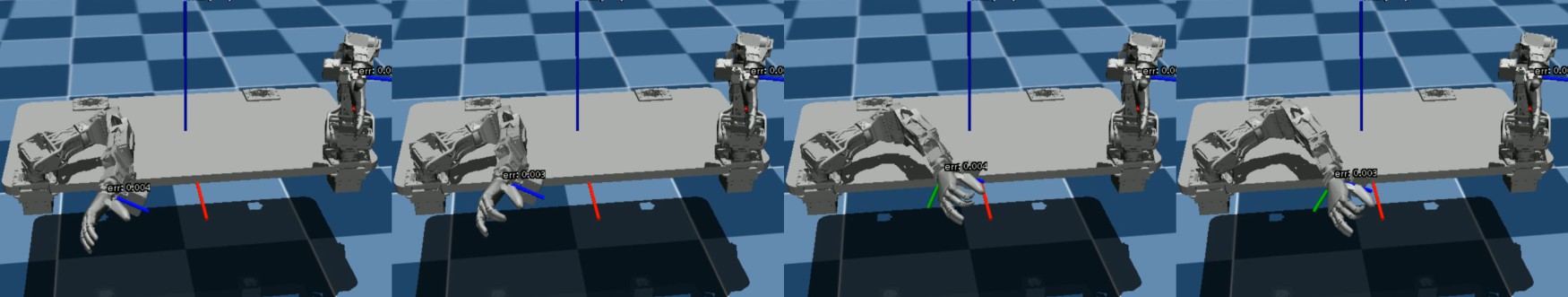}
    \caption{Dual-arm on a table in simulation}
    \label{fig:sim_dual_arm_peg}
    \end{subfigure}
    
    \begin{subfigure}{0.96\columnwidth}
    \centering
    \includegraphics[width=\columnwidth]{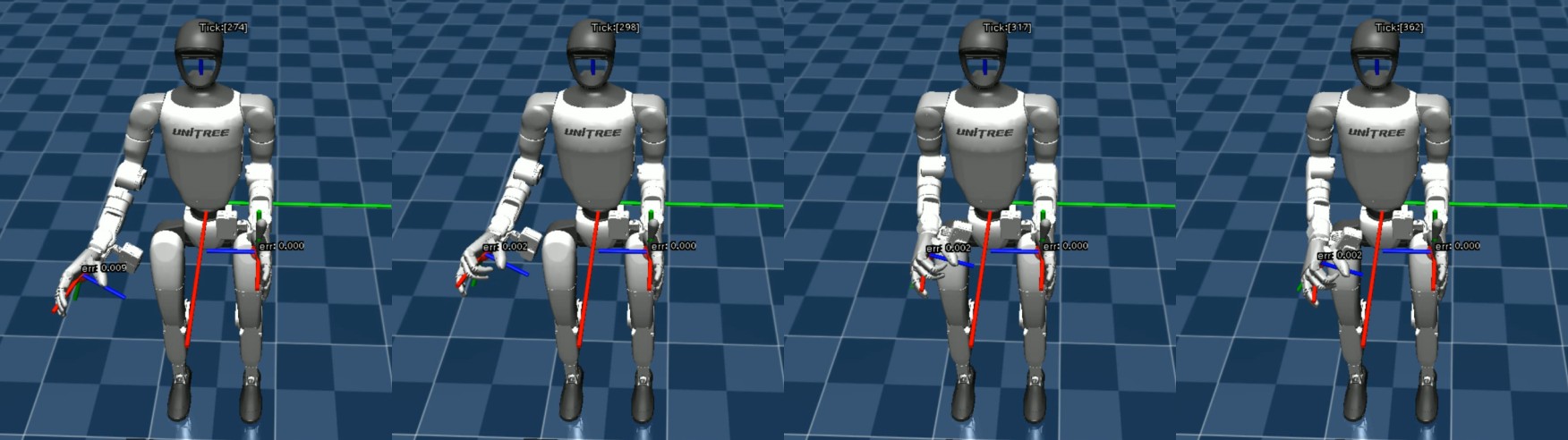}
    \caption{Humanoid robot in simulation}
    \label{fig:sim_g1_peg}
    \end{subfigure} 
\caption{Wrist-trajectory generation and cross-embodiment deployment of data collected with DEX-PHD.} 
\label{fig:demonstration}
\end{figure}

% As shown in Fig.~\ref{fig:human_demo_peg}, a user performs data collection for the assigned tasks via DEX-PHD, where all sensor data are recorded to an internal SD card. 
Fig.~\ref{fig:cam_view_peg} presents the visual data from the wrist-mounted camera, where the wide-angle fisheye lens ensures that the robotic hand, manipulated objects, and the operational environment are all captured simultaneously.

\subsection{Inverse Kinematics}\label{sec:inverse_kinematics}

Each embodiment possesses a unique configuration, characterized by its degrees of freedom (DoFs) and physical dimensions. To replicate identical end effector motions across diverse robotic platforms, we solve the inverse kinematics (IK) problem for each specific embodiment. 

We employ a closed-loop IK solver based on the Pinocchio library \cite{carpentier2019pinocchio} to compute joint trajectories while strictly adhering to joint limits as well as velocity and acceleration bounds. During the motion replay, joint commands are updated at each timestep and are rate-limited to ensure compliance with the mechanical velocity constraints.

To deploy an identical motion sequence across heterogeneous platforms, we apply embodiment-specific initial pose offsets to account for the structural differences in each robot's workspace. Fig.~\ref{fig:trajectory_generation} illustrates the resulting trajectories for various embodiments, demonstrating how these offsets are utilized to maintain the motion within the reachable workspace of each respective robot.

\subsection{Cross-Embodiment Deployment}
Building on the embodiment-specific IK procedure described in Section~\ref{sec:inverse_kinematics}, the estimated wrist trajectories are deployed across robot platforms with different kinematic configurations.

To validate replayability in simulation, we evaluate two embodiments with different kinematics. As shown in Fig.~\ref{fig:sim_dual_arm_peg} and Fig.~\ref{fig:sim_g1_peg}, the peg-in-hole insertion task is replayed on a dual-arm setup and a humanoid robot using the same demonstration sequence. Despite differences in kinematic structures, both embodiments execute consistent task behaviors through embodiment-specific IK solutions and offsets, demonstrating accurate trajectory replay.

% Fig.~\ref{fig:hw_g1_peg} shows further real-world deployment on the physical humanoid robot. Since the peg-in-hole tasks require accurate trajectory tracking, the accuracy is critical to complete the task. The same target task is reproduced on hardware using the identical demonstration, confirming that the proposed pipeline supports accurate trajectory replay beyond simulation.

\subsection{Experimental Setup}
\begin{figure}[t]
\vspace*{5pt}
\centering
    \includegraphics[width=0.7\columnwidth]{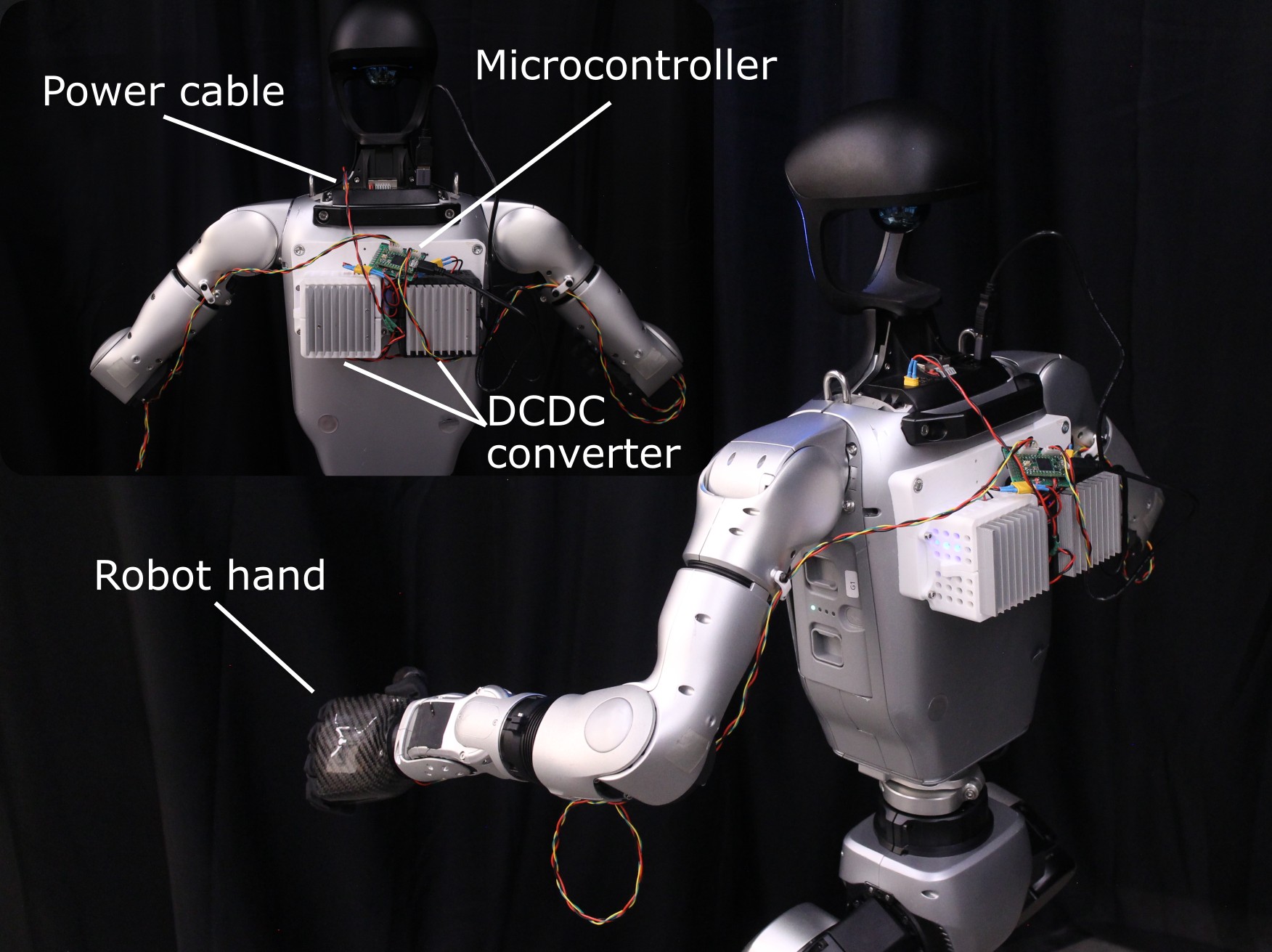}
    \caption{Hardware setup for the humanoid robot.}
    \label{fig:g1_setup}
\end{figure}
Additional components are required to interface and operate the robotic hands on the humanoid platform, as illustrated in Fig.~\ref{fig:g1_setup}. The humanoid robot utilizes the ROS 2 middleware \cite{macenski2022robot}. Similar to the DEX-PHD setup, the robotic hands are controlled by a dedicated microcontroller (Teensy 4.0) running micro-ROS, which communicates with the humanoid's embedded local computer via USB \cite{belsare2023micro}. 

To manage power distribution, since the humanoid robot provides a 54\,V DC supply, two DC-DC step-down converters are mounted on the robot's backplate to regulate the voltage to the required 7.4\,V for the hands. During deployment, only the two arms are actively controlled.

For each task, we collected ten demonstration episodes using DEX-PHD. Prior to physical replay, each episode was evaluated for workspace feasibility and end effector tracking error under the embodiment-specific inverse kinematics solution. Among the feasible episodes, we selected the episode with the lowest end effector tracking error and replayed it once on the physical humanoid robot.

\subsection{Results}

\begin{figure}[!t]
\vspace*{5pt}
\centering
    % ADDED: peg-in-hole figure
    \begin{subfigure}{0.96\columnwidth}
    \centering
    \includegraphics[width=\columnwidth]{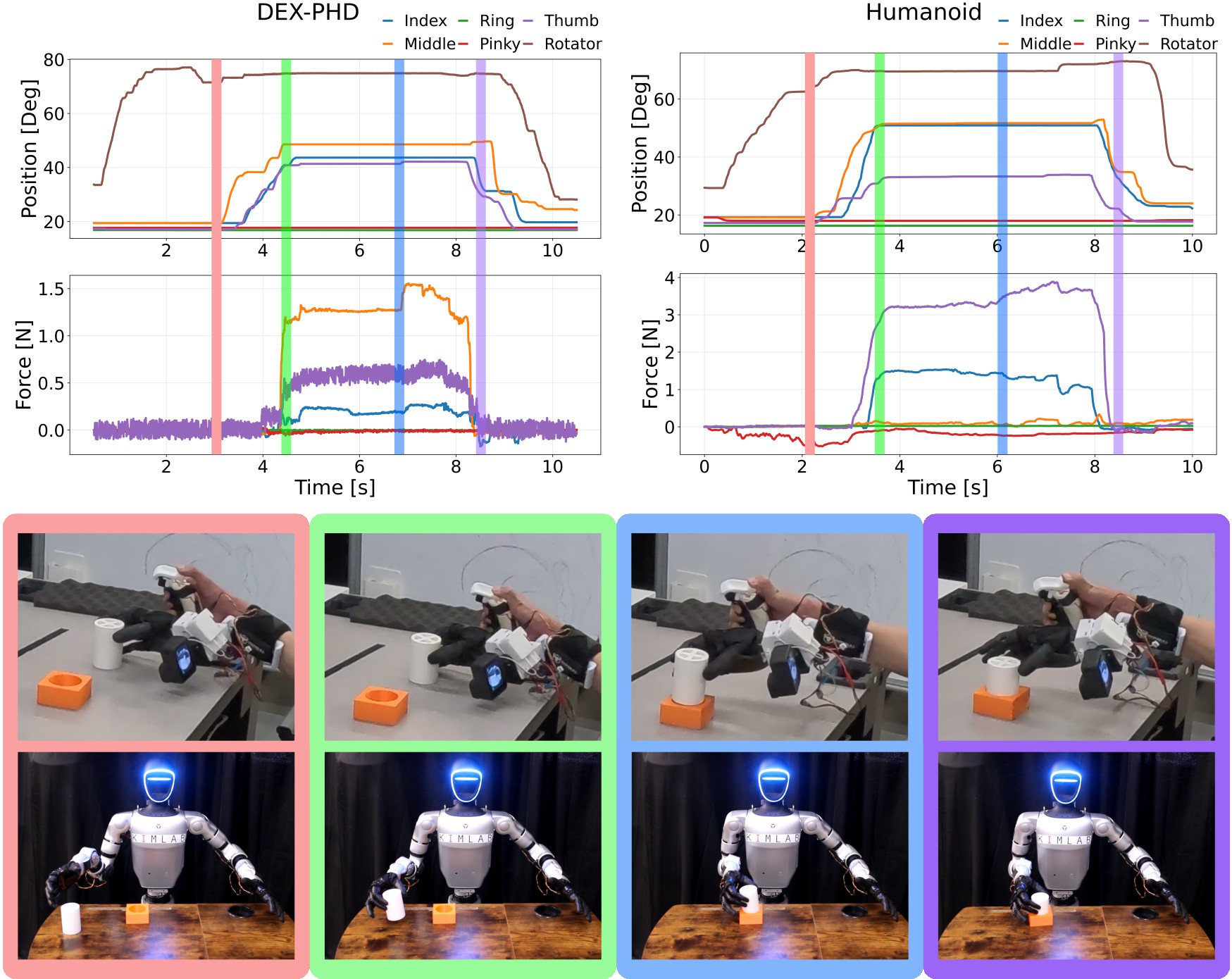}
    \caption{Peg-in-hole}
    \label{fig:force_peg}
    \end{subfigure}

    \begin{subfigure}{0.96\columnwidth}
    \centering
    \includegraphics[width=\columnwidth]{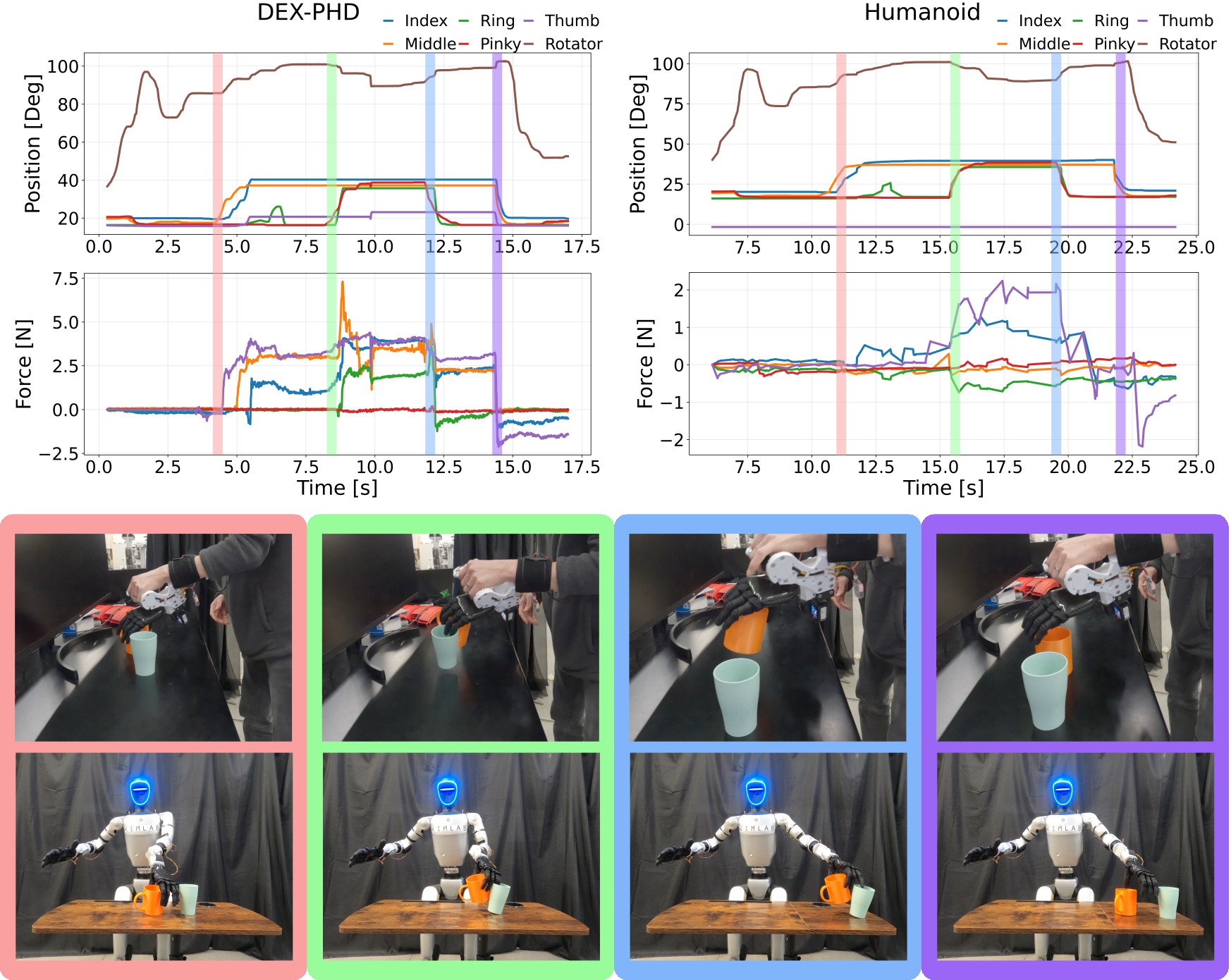}
    \caption{Multi-object grasping}
    \label{fig:force_cups}
    \end{subfigure} 
    
    \begin{subfigure}{0.96\columnwidth}
    \centering
    \includegraphics[width=\columnwidth]{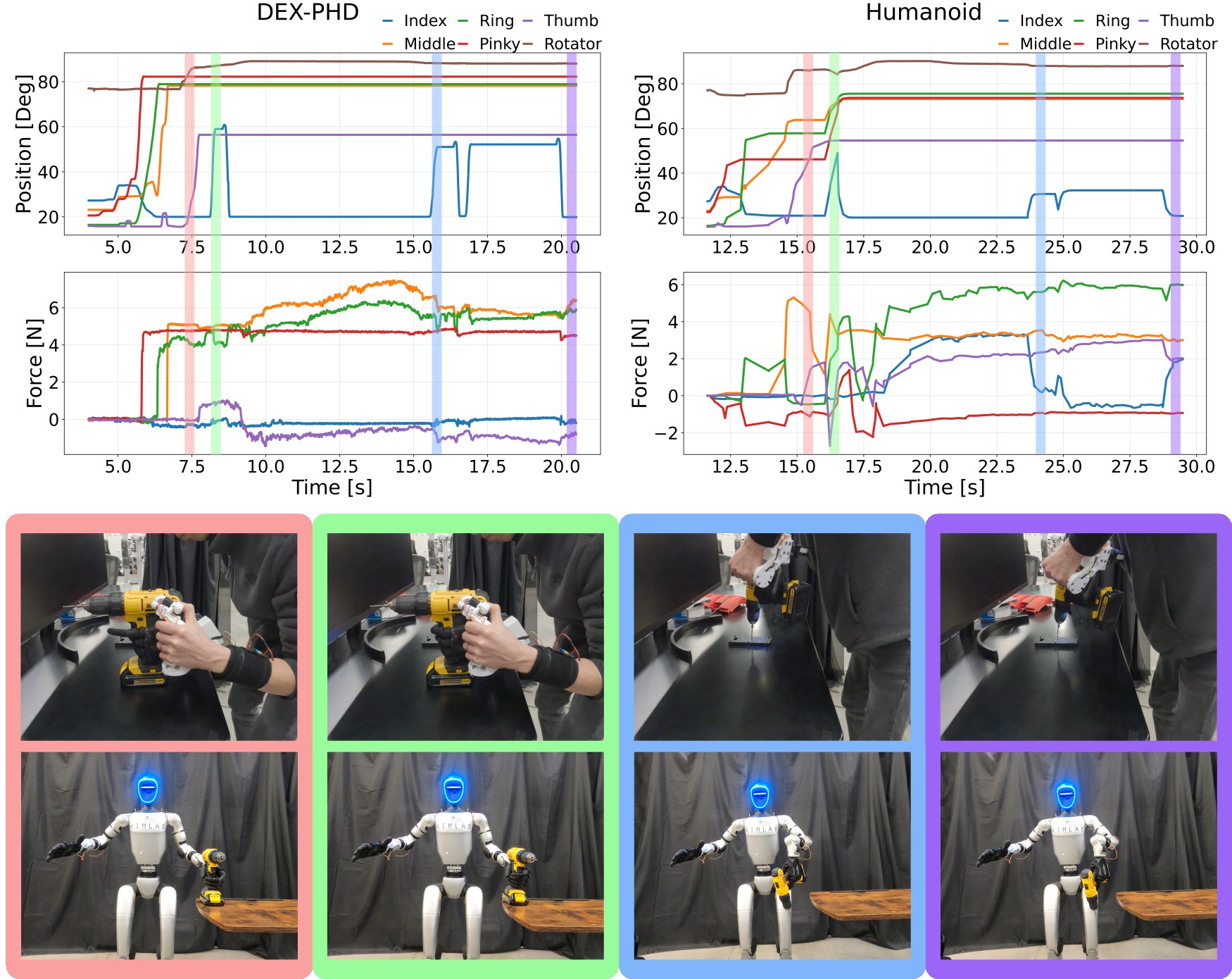}
    \caption{Drilling}
    \label{fig:force_drill}
    \end{subfigure}      
\caption{Force and position trajectories of fingers during tasks.} 
\label{fig:demo_results}
\end{figure}

% In this section, the peg-in-hole, multi-object grasping, and drilling tasks are analyzed to evaluate how the physical robot executes the same tasks using trajectories collected from human demonstrations.
% Fig.~\ref{fig:demo_results} illustrates the joint position and force data of the hand for both tasks during data collection and physical deployment. 

% ADDED: quantitative results table
\begin{table}[t]
\vspace*{5pt}
\centering
\caption{Quantitative analysis}
\label{table:quantitative_analysis}
% \resizebox{\columnwidth}{!}{
\begin{tabular}{lccc}
\hline
Task &
\shortstack{Joint position \\ MAE [deg]}  &
\shortstack{Joint position \\ correlation} &
% All-finger FSR corr. &
\shortstack{Task-relevant \\ contact correlation} \\
\hline
Peg-in-hole &
2.66 &%/ 3.51 &
0.560 &
% -0.072 &
0.197 \\
Multi-grasping &
5.48 &%/ 7.36 &
0.691 &
% 0.091 &
0.691 \\
Drilling &
7.54 &%/ 10.46 &
0.803 &
% 0.091 &
0.761 \\
\hline
\end{tabular}
% }
\end{table}
% ADDED: mention peg-in-hole and quantitative analysis without changing the original sentence above

% We evaluate the replayability of three tasks on the humanoid robot. 
% For each task, we collected ten demonstration episodes using DEX-PHD. Before physical replay, we selected the episode whose wrist trajectory was feasible within the humanoid robot’s workspace and yielded the lowest end effector tracking error under the embodiment-specific inverse kinematics solution. The selected episode was replayed once on the physical humanoid robot for each task. Therefore, the quantitative results characterize one selected demonstration–replay pair per task and do not measure variability across repeated executions or task success rates.

The quantitative results characterize one selected demonstration–replay pair per task and therefore do not measure variability across repeated executions or task success rates.
Fig.~\ref{fig:demo_results} compares the joint position trajectories and fingertip forces recorded during DEX-PHD demonstration and physical humanoid replay, and Table~\ref{table:quantitative_analysis} summarizes the corresponding quantitative results. The DEX-PHD hand trajectories are used as the reference, and the replayed G1 hand trajectories are treated as the measured response after resampling both signals onto a common time axis. 
% Joint position error is reported as mean absolute error (MAE) to quantify absolute tracking deviation, while joint position correlation is used to evaluate whether the replay preserves the temporal shape of the demonstrated finger motions. 
Joint position error is reported as mean absolute error (MAE) to quantify absolute tracking deviation, while joint position correlation and contact correlation evaluate temporal trajectory and contact-pattern similarity, respectively. For both, values closer to 1 indicate better agreement between the DEX-PHD demonstration and physical replay.
The fingertip force signals are analyzed as baseline-relative measurements. For each trial, the initial non-contact interval is used as the reference state, and the corresponding per-channel baseline is subtracted from subsequent measurements. Thus, the plotted values represent changes relative to the initial state rather than independently calibrated absolute force magnitudes. Contact correlation evaluates task-relevant contact-pattern similarity rather than absolute force reproduction. The six FSR values on each finger are aggregated into a single finger-level contact signal to reduce sensitivity to small differences in contact location.

\subsubsection{Peg-in-hole} The peg-in-hole task evaluates replay under constrained insertion, where small differences in peg alignment, contact timing, or hand pose can strongly affect the measured contact profile. As shown in Fig.~\ref{fig:force_peg}, the replayed finger trajectories generally follow the recorded DEX-PHD trajectories. This task produced the lowest joint tracking error among the evaluated tasks, with a mean joint MAE of 2.66$^{\circ}$. The active grasping fingers preserved the temporal structure of the recorded motion, with high joint position correlations for the index, middle, and thumb fingers. The task-relevant contact correlation for peg-in-hole was 0.197. 

% This lower value is expected for constrained insertion because small shifts in the peg-hole alignment or contact onset can substantially change which fingertip sensors are loaded, even when the joint trajectories are replayed accurately.
The low contact correlation indicates that agreement in joint motion does not necessarily imply similar fingertip contact signals during open-loop insertion. Differences in peg–hole alignment and contact onset may contribute to this discrepancy, although their individual effects were not isolated in this evaluation.
Overall, the peg-in-hole result shows that DEX-PHD can reproduce the joint-level finger motion required for constrained insertion, while contact profiles remain sensitive to physical alignment during open-loop replay. 

\subsubsection{Multi-object grasping} The multi-object grasping task requires sequential finger reconfiguration. The first cup is grasped with a tripod grasp, while the second cup is grasped using the remaining fingers and then released in reverse order. As shown in Fig.~\ref{fig:force_cups}, the replayed trajectories preserve the overall timing and ordering of the demonstrated grasp and release events. The task achieved a mean joint MAE of 5.48$^{\circ}$ and a joint position correlation of 0.691. The primary grasping contacts were also reproduced, with a task-relevant contact correlation of 0.691 computed from the index finger and thumb. 

Differences in the secondary contact profiles are mainly attributed to small changes in cup placement and fingertip contact location during replay. Because the contact point can shift across the six FSRs on each finger, the measured baseline-relative force profiles can vary even when the grasp and release events occur in the correct order. Despite these variations, the main contact transitions associated with grasping and releasing the cups remain clearly visible.

\subsubsection{Drilling} 
The drilling task requires coordinated tool grasping, thumb opposition, and independent index-finger actuation of the drill trigger. The complete task, including picking up the drill, fastening a screw, and replacing the tool, was demonstrated using DEX-PHD. During humanoid replay, however, the analysis focused on the in-air drilling motion because the drill weight and whole-body posture constraints made the full fastening sequence difficult to reproduce reliably on the humanoid platform. As shown in Fig.~\ref{fig:force_drill}, the replayed hand motion preserves the main temporal structure of the demonstrated grasp and trigger-actuation sequence. Although drilling produced the largest joint MAE among the tasks at 7.54$^{\circ}$, it also showed the highest joint position correlation of 0.803, indicating that the overall motion pattern was largely retained despite pose shifts of the tool. The task-relevant contact correlation was 0.761, computed from the middle and ring fingers that maintained the primary power grasp on the drill handle. 

Variations in object positioning, tool weight, manipulator sagging, and torso orientation affected the drill pose during replay, so contact was not always applied to the same fingertip regions or with the same magnitude. As a result, the baseline-relative force profiles differ in timing and amplitude, while still preserving the main contact transitions associated with the power grasp and trigger-actuation sequence. These results indicate that DEX-PHD can replay coordinated per-finger command patterns for tool-use tasks while revealing how contact signals are affected by physical replay conditions.

\section{Conclusion}
% This paper presented DEX-PHD, a portable handheld device for collecting multimodal demonstrations with dexterous robotic hands. By physically integrating a dexterous robot hand into the handheld device, DEX-PHD enables continuous per-finger joint-level command generation while synchronously recording wrist-camera video, measured finger states, and fingertip force signals.

This paper presented DEX-PHD, a portable handheld device for collecting multimodal demonstrations with dexterous robotic hands. By physically integrating a dexterous robot hand into the handheld device, DEX-PHD enables continuous per-finger joint-level command generation while synchronously recording wrist-camera, hand states, and fingertip force signals. Its battery-powered and onboard logging architecture further enables untethered data collection across diverse indoor and outdoor environments without requiring a fixed robot platform or external PC.

Across peg-in-hole insertion, multi-object grasping, and drilling tasks, DEX-PHD collected replayable demonstrations that preserved the main temporal structure of finger motions during physical replay. The quantitative results showed task-dependent joint tracking errors, trajectory-shape similarity, and task-relevant contact-pattern agreement, suggesting that synchronized hand-state and contact signals can provide useful evaluation signals for contact-rich manipulation.

Several limitations remain. The current replay experiments are primarily open-loop, so contact timing and fingertip force profiles can vary with object pose, alignment, and whole-body motion errors. Therefore, fingertip force signals are analyzed as baseline-relative contact patterns rather than independently calibrated absolute force measurements. In addition, the current interface is designed for the six actuated degrees of freedom of the robotic hand. Future work will extend the mounting structure, input mechanisms, and command mapping to support other dexterous robotic hands, including hands with additional independently actuated joints.
% Future work will investigate closed-loop replay and policy learning using multimodal data collected by DEX-PHD.


% %%%%%%%%%%%%%%%%%%%%%%%%%%%%%%%%%%%%%%%%%%%%%%%%%%%%%%%%%%%%%%%%%%%%%%%%%%%%%%%%

% %%%%%%%%%%%%%%%%%%%%%%%%%%%%%%%%%%%%%%%%%%%%%%%%%%%%%%%%%%%%%%%%%%%%%%%%%%%%%%%%



% \section*{ACKNOWLEDGMENT}
% This work was partially supported by Toyota Research Institute (TRI).
\section*{ACKNOWLEDGMENT}
This work was partially supported by Toyota Research Institute (TRI).
% The preferred spelling of the word ÒacknowledgmentÓ in America is without an ÒeÓ after the ÒgÓ. Avoid the stilted expression, ÒOne of us (R. B. G.) thanks . . .Ó  Instead, try ÒR. B. G. thanksÓ. Put sponsor acknowledgments in the unnumbered footnote on the first page.

% %%%%%%%%%%%%%%%%%%%%%%%%%%%%%%%%%%%%%%%%%%%%%%%%%%%%%%%%%%%%%%%%%%%%%%%%%%%%%%%%

% References are important to the reader; therefore, each citation must be complete and correct. If at all possible, references should be commonly available publications.

\bibliography{ref}

@INPROCEEDINGS{11128816,
  author={Zhao, Jialiang and Kuppuswamy, Naveen and Feng, Siyuan and Burchfiel, Benjamin and Adelson, Edward},
  booktitle={2025 IEEE International Conference on Robotics and Automation (ICRA)}, 
  title={PolyTouch: A Robust Multi-Modal Tactile Sensor for Contact-Rich Manipulation Using Tactile-Diffusion Policies}, 
  year={2025},
  volume={},
  number={},
  pages={104-110},
  doi={10.1109/ICRA55743.2025.11128816}}

@article{sankar2025natural,
  title={A natural biomimetic prosthetic hand with neuromorphic tactile sensing for precise and compliant grasping},
  author={Sankar, Sriramana and Cheng, Wen-Yu and Zhang, Jinghua and Slepyan, Ariel and Iskarous, Mark M and Greene, Rebecca J and DeBrabander, Rene and Chen, Junjun and Gupta, Arnav and Thakor, Nitish V},
  journal={Science Advances},
  volume={11},
  number={10},
  year={2025},
  publisher={American Association for the Advancement of Science}
}

@article{huang2025dih,
  title={Dih-tele: Dexterous in-hand teleoperation framework for learning multiobjects manipulation with tactile sensing},
  author={Huang, Junda and Chen, Kai and Zhou, Jianshu and Lin, Xingyu and Abbeel, Pieter and Dou, Qi and Liu, Yunhui},
  journal={IEEE/ASME Transactions on Mechatronics},
  year={2025},
  publisher={IEEE}
}

@inproceedings{qin2023anyteleop,
  title     = {AnyTeleop: A General Vision-Based Dexterous Robot Arm-Hand Teleoperation System},
  author    = {Qin, Yuzhe and Yang, Wei and Huang, Binghao and Van Wyk, Karl and Su, Hao and Wang, Xiaolong and Chao, Yu-Wei and Fox, Dieter},
  booktitle = {Robotics: Science and Systems (RSS)},
  year      = {2023}
}

@inproceedings{zhao2023learning,
  title = {Learning fine-grained bimanual manipulation with low-cost hardware},
  author = {Zhao, Tony Z and Kumar, Vikash and Levine, Sergey and Finn, Chelsea},
  booktitle = {Robotics: Science and Systems (RSS)},
  year = {2023}
}

@INPROCEEDINGS{Wu2024GELLO,
  author={Wu, Philipp and Shentu, Yide and Yi, Zhongke and Lin, Xingyu and Abbeel, Pieter},
  booktitle={2024 IEEE/RSJ International Conference on Intelligent Robots and Systems (IROS)}, 
  title={GELLO: A General, Low-Cost, and Intuitive Teleoperation Framework for Robot Manipulators}, 
  year={2024},
  volume={},
  number={},
  pages={12156-12163},
  doi={10.1109/IROS58592.2024.10801581}}

@inproceedings{
yang2024ace,
title={{ACE}: A Cross-platform and visual-Exoskeletons System for Low-Cost Dexterous Teleoperation},
author={Shiqi Yang and Minghuan Liu and Yuzhe Qin and Runyu Ding and Jialong Li and Xuxin Cheng and Ruihan Yang and Sha Yi and Xiaolong Wang},
booktitle={8th Annual Conference on Robot Learning},
year={2024}
}

@INPROCEEDINGS{tao2025dexwild,
      title={DexWild: Dexterous Human Interactions for In-the-Wild Robot Policies},
      author={Tao, Tony and Srirama, Mohan Kumar and Liu, Jason Jingzhou and Shaw, Kenneth and Pathak, Deepak},
      booktitle={Robotics: Science and Systems (RSS)},
      year={2025}}

@inproceedings{chi2024universal,
	title={Universal Manipulation Interface: In-The-Wild Robot Teaching Without In-The-Wild Robots},
	author={Chi, Cheng and Xu, Zhenjia and Pan, Chuer and Cousineau, Eric and Burchfiel, Benjamin and Feng, Siyuan and Tedrake, Russ and Song, Shuran},
	booktitle={Robotics: Science and Systems (RSS)},
	year={2024}
}

@ARTICLE{seo2025legato,
  author={Seo, Mingyo and Park, H. Andy and Yuan, Shenli and Zhu, Yuke and Sentis, Luis},
  journal={IEEE Robotics and Automation Letters}, 
  title={LEGATO: Cross-Embodiment Imitation Using a Grasping Tool}, 
  year={2025},
  volume={10},
  number={3},
  pages={2854-2861},
  doi={10.1109/LRA.2025.3535182}}

@inproceedings{liu2025forcemimic,
  title={Forcemimic: Force-centric imitation learning with force-motion capture system for contact-rich manipulation},
  author={Liu, Wenhai and Wang, Junbo and Wang, Yiming and Wang, Weiming and Lu, Cewu},
  booktitle={2025 IEEE International Conference on Robotics and Automation (ICRA)},
  pages={1105--1112},
  year={2025}
}

@inproceedings{
iyer2024open,
title={{OPEN} {TEACH}: A Versatile Teleoperation System for Robotic Manipulation},
author={Aadhithya Iyer and Zhuoran Peng and Yinlong Dai and Irmak Guzey and Siddhant Haldar and Soumith Chintala and Lerrel Pinto},
booktitle={8th Annual Conference on Robot Learning},
year={2024}
}

@INPROCEEDINGS{kareer2025egomimic,
  author={Kareer, Simar and Patel, Dhruv and Punamiya, Ryan and Mathur, Pranay and Cheng, Shuo and Wang, Chen and Hoffman, Judy and Xu, Danfei},
  booktitle={2025 IEEE International Conference on Robotics and Automation (ICRA)}, 
  title={EgoMimic: Scaling Imitation Learning via Egocentric Video}, 
  year={2025},
  volume={},
  number={},
  pages={13226-13233},
  doi={10.1109/ICRA55743.2025.11127989}}

@INPROCEEDINGS{11128180,
  author={Lin, Toru and Zhang, Yu and Li, Qiyang and Qi, Haozhi and Yi, Brent and Levine, Sergey and Malik, Jitendra},
  booktitle={2025 IEEE International Conference on Robotics and Automation (ICRA)}, 
  title={Learning Visuotactile Skills With Two Multifingered Hands}, 
  year={2025},
  volume={},
  number={},
  pages={5637-5643},
  doi={10.1109/ICRA55743.2025.11128180}}

@inproceedings{
cheng2024opentelevision,
title={Open-TeleVision: Teleoperation with Immersive Active Visual Feedback},
author={Xuxin Cheng and Jialong Li and Shiqi Yang and Ge Yang and Xiaolong Wang},
booktitle={8th Annual Conference on Robot Learning},
year={2024}
}

@inproceedings{
zhu2025touch,
title={Touch in the Wild: Learning Fine-Grained Manipulation with a Portable Visuo-Tactile Gripper},
author={Xinyue Zhu and Binghao Huang and Yunzhu Li},
booktitle={The Thirty-ninth Annual Conference on Neural Information Processing Systems},
year={2025}
}

@inproceedings{
xu2025dexumi,
title={Dex{UMI}: Using Human Hand as the Universal Manipulation Interface for Dexterous Manipulation},
author={Mengda Xu and Han Zhang and Yifan Hou and Zhenjia Xu and Linxi Fan and Manuela Veloso and Shuran Song},
booktitle={9th Annual Conference on Robot Learning},
year={2025}
}

@INPROCEEDINGS{sanches2023scalable,
  author={Sanches, Felipe and Gao, Geng and Elangovan, Nathan and Godoy, Ricardo V. and Chapman, Jayden and Wang, Ke and Jarvis, Patrick and Liarokapis, Minas},
  booktitle={2023 IEEE/RSJ International Conference on Intelligent Robots and Systems (IROS)}, 
  title={Scalable, Intuitive Human to Robot Skill Transfer with Wearable Human Machine Interfaces: On Complex, Dexterous Tasks}, 
  year={2023},
  volume={},
  number={},
  pages={6318-6325},
  doi={10.1109/IROS55552.2023.10341661}}

@inproceedings{wang2024dexcap,
  author={Chen Wang and Haochen Shi and Weizhuo Wang and Ruohan Zhang and Li Fei-Fei and C. Karen Liu},
  title={DexCap: Scalable and Portable Mocap Data Collection System for Dexterous Manipulation},
  year={2024},
  booktitle={Robotics: Science and Systems (RSS)}
}

@misc{Psyonic,
  author       = {{Psyonic}},
  title        = {Ability Hand Interface Control Document},
  howpublished = {\url{https://github.com/psyonicinc/ability-hand-api/blob/master/Documentation/ABILITY-HAND-ICD.pdf}},
  note         = {Accessed: 2026-06-25}
}

@misc{unitree_g1,
  author       = {{Unitree Robotics}},
  title        = {Unitree G1: Humanoid Agent AI Avatar},
  howpublished = {\url{https://www.unitree.com/g1}},
  note         = {Accessed: 2026-06-25}
}

@article{macenski2022robot,
  title={Robot operating system 2: Design, architecture, and uses in the wild},
  author={Macenski, Steven and Foote, Tully and Gerkey, Brian and Lalancette, Chris and Woodall, William},
  journal={Science Robotics},
  volume={7},
  number={66},
  year={2022},
  publisher={American Association for the Advancement of Science}
}

@incollection{belsare2023micro,
  title={Micro-ros},
  author={Belsare, Kaiwalya and Rodriguez, Antonio Cuadros and S{\'a}nchez, Pablo Garrido and Hierro, Juanjo and Ko{\l}con, Tomasz and Lange, Ralph and L{\"u}tkebohle, Ingo and Malki, Alexandre and Losa, Jaime Martin and Melendez, Francisco and others},
  booktitle={Robot Operating System (ROS) The Complete Reference (Volume 7)},
  pages={3--55},
  year={2023},
  publisher={Springer}
}

@inproceedings{carpentier2019pinocchio,
   title={The Pinocchio C++ library -- A fast and flexible implementation of rigid body dynamics algorithms and their analytical derivatives},
   author={Carpentier, Justin and Saurel, Guilhem and Buondonno, Gabriele and Mirabel, Joseph and Lamiraux, Florent and Stasse, Olivier and Mansard, Nicolas},
   booktitle={IEEE International Symposium on System Integration (SII)},
   year={2019}
}

@inproceedings{
fang2025dexop,
title={{DEXOP}: Hand Exoskeleton System for Teaching Robot Dexterous Manipulation In-The-Wild},
author={Hao-Shu Fang and Branden Romero and Arthur Hu and Edward H Adelson and Pulkit Agrawal},
booktitle={3rd RSS Workshop on Dexterous Manipulation: Learning and Control with Diverse Data},
year={2025},

}

@ARTICLE{Tang2026MFE,
  author={Tang, Ziyuan and Guo, Yitian and Xiao, Chenxi},
  journal={IEEE Robotics and Automation Letters}, 
  title={MFE: A Multimodal Hand Exoskeleton With Interactive Force, Pressure and Thermo-Haptic Feedback}, 
  year={2026},
  volume={11},
  number={3},
  pages={3756-3763},
  doi={10.1109/LRA.2026.3662616}}

@inproceedings{xue2025reactive,
  title     = {Reactive Diffusion Policy: Slow-Fast Visual-Tactile Policy Learning for Contact-Rich Manipulation},
  author    = {Xue, Han and Ren, Jieji and Chen, Wendi and Zhang, Gu and Fang, Yuan and Gu, Guoying and Xu, Huazhe and Lu, Cewu},
  booktitle = {Robotics: Science and Systems (RSS)},
  year      = {2025}
}

@inproceedings{
huang2025vtrefine,
title={{VT}-Refine: Learning Bimanual Assembly with Visuo-Tactile Feedback via Simulation Fine-Tuning},
author={Binghao Huang and Jie Xu and Iretiayo Akinola and Wei Yang and Balakumar Sundaralingam and Rowland O'Flaherty and Dieter Fox and Xiaolong Wang and Arsalan Mousavian and Yu-Wei Chao and Yunzhu Li},
booktitle={9th Annual Conference on Robot Learning},
year={2025}
}

@inproceedings{chen2026multimodal,
  title={Multi-Modal Manipulation via Multi-Modal Policy Consensus},
  author={Chen, Haonan and Xu, Jiaming and Chen, Hongyu and Hong, Kaiwen and Huang, Binghao and Liu, Chaoqi and Mao, Jiayuan and Li, Yunzhu and Du, Yilun and Driggs-Campbell, Katherine},
  booktitle={2026 IEEE International Conference on Robotics and Automation (ICRA)},
  year={2026},
  note={to appear, arXiv:2509.23468}
}

@INPROCEEDINGS{Zhang2025DOGlove, 
    AUTHOR    = {Han Zhang AND Songbo Hu AND Zhecheng Yuan AND Huazhe Xu}, 
    TITLE     = {{DOGlove: Dexterous Manipulation with a Low-Cost Open-Source Haptic Force Feedback Glove}}, 
    BOOKTITLE = {Robotics: Science and Systems}, 
    YEAR      = {2025}
}
\bibliographystyle{IEEEtran}

\end{document}